\RequirePackage{fix-cm}
\RequirePackage{amsmath}
\documentclass[twocolumn]{svjour3}          % twocolumn
\smartqed  % flush right qed marks, e.g. at end of proof
\usepackage{epsfig}
\usepackage{amssymb}
\usepackage{float}
\usepackage{subcaption}
\usepackage{graphicx}
\usepackage{cite}
\usepackage{algorithm,algorithmic}
\usepackage[usenames, dvipsnames]{color}
\usepackage{multirow}
\begin{document}

\title{Studying the Plasticity in Deep Convolutional Neural Networks using Random Pruning
}
% \subtitle{Do you have a subtitle?\\ If so, write it here}


\author{\small{Deepak Mittal} \and
        \small{Shweta Bhardwaj} \and
        \small{Mitesh M. Khapra}\and
        \small{Balaraman Ravindran} %etc.
}

\institute{D. Mittal, S. Bhardwaj, M. M. Khapra and B. Ravindran \at
              Department of Computer Science and Engineering \\
              Robert Bosch Centre for Data Science and AI (RBC-DSAI)\\
              Indian Institute of Technology Madras, Chennai, India\\
              \email{\{deepak, cs16s003, miteshk, ravi\}@cse.iitm.ac.in}         }

\date{Received: date / Accepted: date}

\maketitle

\begin{abstract}
Recently there has been a lot of work on pruning filters from deep convolutional neural networks (CNNs) with the intention of reducing computations. The key idea is to rank the filters based on a certain criterion (say, $l_1$-norm, average percentage of zeros, etc) and retain only the top ranked filters. Once the low scoring filters are pruned away the remainder of the network is fine tuned and is shown to give performance comparable to the original unpruned network. In this work, we report experiments which suggest that the comparable performance of the pruned network is not due to the specific criterion chosen but due to the inherent plasticity of deep neural networks which allows them to recover from the loss of pruned filters once the rest of the filters are fine-tuned. Specifically, we show counter-intuitive results wherein by randomly pruning $25$-$50$\% filters from deep CNNs we are able to obtain the same performance as obtained by using state-of-the-art pruning methods. We empirically validate our claims by doing an exhaustive evaluation with VGG-16 and ResNet-$50$. Further, we also evaluate a real world scenario where a CNN trained on all 1000 ImageNet classes needs to be tested on only a small set of classes at test time (say, only animals). We create a new benchmark dataset from ImageNet to evaluate such class specific pruning and show that even here a random pruning strategy gives close to state-of-the-art performance. Lastly, unlike existing approaches which mainly focus on the task of image classification, in this work we also report results on object detection and image segmentation. We show that using a simple random pruning strategy we can achieve significant speed up in object detection (74$\%$ improvement in fps) while retaining the same accuracy as that of the original Faster RCNN model. Similarly, we show that the performance of a pruned Segmentation Network (SegNet) is actually very similar to that of the original unpruned SegNet.
\end{abstract}

%%%%%%%%% INTRODUCTION %%%%%%%%%%%%%%
\section{Introduction}
Over the past few years, deep convolutional neural networks (CNNs) have been very successful in a wide range of computer vision tasks such as image classification  \cite{inception,xception,imagenet-cnn} , object detection \cite{rcnn,fastrcnn,f-rcnn,yolo,ssd} and image segmentation \cite{segnet,segmentation}. In general, with each passing year, these networks are becoming deeper and deeper with a corresponding increase in the performance \cite{resnet,densenet,highwaynet}. However, this increase in performance is accompanied by an increase in the number of parameters and computations. This makes it difficult to port these models on embedded and mobile devices where storage, computation and power are limited. In such cases, it is crucial to have small, computationally efficient models which can achieve performance at par or close to large networks. This practical requirement has led to an increasing interest in model compression where the aim is to either (i) design efficient small networks \cite{SqueezeNet,mobilenets} or (ii) efficiently prune weights from existing deep networks \cite{learn-weights-connections,structured-sparsity,deepcompression} or (iii) efficiently prune filters from deep convolutional networks \cite{l1norm,thinet,efficientInference,apoz} or (iv) replace expensive floating point weights by binary or quantized weights \cite{binary, incremental-quantization, xnor,deepcompression} or (v) guide the training of a smaller network using a larger (teacher) network \cite{mimic-learning-network-distillation,distillation}.

In this work, we focus on pruning filters from deep convolutional neural networks. The filters in the convolution layers typically account for fewer parameters than the fully connected layers (the ratio is 10:90 for VGG-16 \cite{l1norm}), but they account for most of the floating point operations done by the model (99\% for VGG-16 \cite{l1norm}). Hence reducing the number of filters effectively reduces the computation (and thus power) requirements of the model. All existing works on filter pruning \cite{l1norm,thinet,efficientInference,apoz} follow a very similar recipe. The filters are first ranked based on a specific criterion such as, $l_1$-norm \cite{l1norm} or percentage of zeros in the filter \cite{apoz}. The scoring criterion essentially determines the importance of the filter for the end task, typically image classification \cite{imagenet-cnn}. Only the top-$m$ ranked filters are retained and the resulting pruned network is then fine tuned. It is observed that when pruning up to 50\% of the filters using different proposed criteria, the pruned network almost recovers the original performance after fine-tuning. The claim is that this recovery is due to soundness of the criterion chosen for pruning. However, in this work we argue that this recovery is not due the specific pruning criterion but due to the inherent plasticity of deep CNNs. Specifically, we show that even if we prune filters randomly we can match the performance of state-of-the-art pruning methods.

To effectively prove our point, it is crucial that we look at factors/measures other than the final performance of the pruned model. To do so we draw an analogy with the human brain and observe that the process of pruning filters from a deep CNN is akin to causing damage to certain portions of the brain. It is known that the human brain has a high plasticity and over the time can recover from such damages with appropriate  treatment \cite{neuroplasticity-1}. In our case, the process of fine-tuning would be akin to such post-damage (post-pruning) treatment. 
If the injury damages only redundant or unimportant portions of the brain then the recovery should be completed quickly and with minimal treatment. Similarly, we could argue that if the pruning criteria is indeed good and prunes away only unimportant filters then (i) the performance of the model should not drop much (ii) the model should be able to regain its full performance after fine-tuning (iii) this recovery should be fast (i.e, with fewer iterations of fine tuning) and (iv) the quantum of data used for fine-tuning 
%(akin to quantum of treatment) 
should be less. None of the existing works on filter pruning do a thorough  comparison w.r.t. these factors. We not only consider these factors but also present counter-intuitive results which show that a random pruning criteria is comparable to state-of-the-art pruning methods on all these factors. Note that we are not claiming that we can always recover the full performance of the unpruned network. For example, it should be obvious that in the degenerate case if 90\% of the filters are pruned then it would be almost impossible to recover. %or if 50\% of the filters are pruned then the recovery would be harder (and less) as compared to when 25\% of the filters are pruned.
The claim being made is that, at different pruning levels (25\%, 50\% or 75\%) a random pruning strategy is not much worse than of state-of-the-art pruning strategies.   

To further prove our point, we wanted to check if such recovery from pruning is task agnostic. In other words, in addition to showing that a network trained for image classification (\textit{task1}) can be pruned efficiently, we also show that same can be done with a network trained for object detection (\textit{task2}). Here again, we show that a random pruning strategy works at par with state-of-the-art pruning methods. Stretching this idea further and continuing the above analogy, we note that once the brain recovers from such damages, it is desirable that in addition to recovering its performance on the tasks that it was good at before the injury, it should also be able to do well on newer tasks. In our case, the corresponding situation would be to take a network pruned and fine-tuned for image classification (\textit{old task}) and plug it into a model for object detection (\textit{new task}). Specifically, we show that when we plug a randomly pruned and fine tuned VGG-16 network into a Faster RCNN model we can get the same performance on object detection as obtained by plugging (i) the original unpruned network or (ii) a network pruned using a state-of-the-art pruning method. This once again hints at the inherent plasticity of deep CNNs which allows them to recover (up to a certain level) irrespective of the pruning strategy.

Finally, we consider the case of class specific pruning which has not been studied in the literature. We note that in many real world scenarios, it is possible that while we have trained an image classification network on a large dataset containing many classes, at test time we may be interested in only a few classes. A case in point, is the task of object detection using the Pascal VOC dataset \cite{pascal-voc-2007}. RCNN and its variants \cite{rcnn,fastrcnn,f-rcnn} use as a sub-component an image classification model trained on all the 1000 ImageNet classes. We hypothesize that this is an overkill and instead create a class specific benchmark dataset from ImageNet which contains only those 52 classes which correspond to the 20 classes in Pascal VOC. Ideally, one would expect that a network trained, pruned and fine-tuned only for these 52 classes when plugged into faster RCNN should do better than a network trained, pruned and fine-tuned on a random set of 52 classes (which are very different from the classes in Pascal VOC). However, we observe that irrespective of which of these networks is plugged into Faster RCNN the final performance after fine-tuning is the same, once again showing the ability to recover from unfavorable situations.      

To the best of our knowledge, this is a first of its kind work on pruning filters which:

\begin{enumerate}
\item Proposes that while assessing the performance of a pruning method, we should consider factors such as amount of damage (drop in performance before fine-tuning), amount of recovery (performance after fine-tuning), speed of recovery and quantum of data required for recovery.
\item Performs extensive evaluation using two image classification networks (VGG-16 and ResNet) and shows that a random pruning strategy gives comparable performance to that of state-of-the-art pruning strategies w.r.t. all the above factors. 
\item Shows that such behavior is task agnostic and a random pruning strategy works well even for the task of object detection and image segmentation. Specifically, we show that by randomly pruning filters from an object detection model we can get a 74$\%$ improvement in fps while maintaining almost the same accuracy (1\% drop) as the original unpruned network. Further, in the case of SegNet the performance of the pruned network is actually better than the unpruned network in some cases.  
\item Shows that pruned networks can adapt with ease to newer tasks.
\item Proposes a new benchmark for evaluating class specific pruning.

\end{enumerate}

\section{Related Work}
In this section, we review existing work on making deep convolutional neural networks efficient w.r.t. their memory and computation requirements while not compromising much on the accuracy. These approaches can be broadly classified into the following categories (i) pruning unimportant weights (ii) low rank factorization (iii) knowledge distillation (iv) designing compact networks from scratch or (v) using binary or quantized weights and (vi) pruning unimportant filters. Below, we first quickly review the related work for the first five categories listed above and then discuss approaches on pruning filters which is the main focus of our work.

Optimal brain damage \cite{obd} and optimal brain surgery \cite{brain-surgery} are two examples of approaches which prune the unimportant weights in the network. A weight is considered unimportant if the output is not very sensitive to this weight. They show that pruning such weights leads to minimal drop in the overall performance of the network. However, these methods are computationally expensive as they require the computation of the Hessian (second order derivative). Another approach is to use low rank factorization of the weight tensor/matrices to reduce the computations \cite{approx5,approx4,approx2,approx1,approx3}. For example, instead of directly multiplying a high dimensional weight tensor $W$ with the input tensor $I$, we could first compute a low rank approximation of $W = U\Sigma V$ where the dimensions of $U$, $\Sigma$ and $V$ are much smaller than the dimensions of $W$. This essentially boils down to decomposing the larger matrix multiplication operation into smaller operations. Also, the low rank approximation ensures that only the important information in the weight matrix is retained. Alternately, researchers have also explored designing compact networks from scratch which have fewer number of layers and/or parameters and/or computations \cite{SqueezeNet}. There are also some approaches which quantize \cite{deepcompression} or binarize \cite{xnor,binary} the weights of a network to reduce both memory footprint and computation time. Another line of work focuses on transferring the knowledge from bigger trained network (or ensemble of networks) to smaller (thin) network \cite{mimic-learning-network-distillation,distillation}.\\
\null \qquad The main focus of our work is on pruning filters from deep CNNs with the intention of reducing computations. As mentioned earlier, while the convolution filters do not account for a large number of parameters, they account for almost all the computations that happen in the network. Here, the idea is to rank the filters using a $scoring\_function$ and then retain only the top scoring functions. For example, in \cite{l1norm}, the authors have used the $l_1$-norm of the filters to rank their importance. The argument is that filters having a lower $l_{1}$-norm will produce smaller activation values which will contribute minimally to the output of that layer. Alternately in \cite{entropy}, authors have proposed entropy as a measure to calculate the importance of a filter. If a filter has high entropy than the filter is considered more informative and hence more important. On the other hand, \cite{apoz} calculate the average percentage of zeros in the corresponding activation maps of filters and hypothesize that filters having more average percentage of zeros in their activation are less important. In \cite{efficientInference} authors have used Taylor series expansion that approximates the change in cost function caused by pruning filters. Unlike \cite{obd}, this method uses information from first derivative only. Another work on pruning filters \cite{thinet} proposes that instead of pruning filters based on current layer's statistics, they should be pruned based on next layer's statistics. Essentially the idea of \cite{thinet} is to look at the activation map of layer $i$+$1$ and prune out the channel which will give the minimum change in output on its removal and its corresponding filter in layer $i$. In \cite{accelerating} authors have proposed a similar idea to \cite{thinet} but instead of removing the filters one by one they proposed to use LASSO regression. Lastly, in \cite{anwar} authors has used particle filtering to prune out the filters. 

\section{Methodology}
In this section, we first formally define the problem of filter-pruning and give a generic algorithm for pruning filters using any appropriate $scoring\_function$. We then discuss existing scoring functions along with some new variants that we propose.

\subsection{Problem Statement}
Suppose there are $K$ convolutional layers in a CNN and the $k^{th}$ layer contains $n_k$ filters. We use $F_{ki}$ to denote the $i^{th}$ filter in the $k^{th}$ layer. Each such filter is a three dimensional tensor, $F_{ki} \in \mathbb{R}^{i_k \times w_{ki} \times h_{ki}}$ where $i_k$ is the number of input channels for layer $k$ and $w_{ki}, h_{ki}$ are the width and height of the $i^{th}$ filter in the $k^{th}$ layer. Our goal is to rank all the filters in layer $k$, $\{F_{k1}, F_{k2}, ..., F_{ki}\}$ and then retain the top-$m_k$ filters where $m_k (< n_k)$ is a hyperparameter which indicates the desired pruning (For example, based on available computation resources, if we want to reduce the number of computations in this layer by half then we can set $m_k = \frac{n_k}{2})$. Let the original output of layer $k$ be denoted by $O^k \in \mathbb{R}^{n_k \times w^{o}_{k} \times h^{o}_{k}}$ where $w^{o}_{k}, h^{o}_{k}$ are the width and height and $n_k$ is the number of channels which is the same as the number of filters. After pruning and retaining only top-$m_k$ filters the size of the output will be reduced to $m_k \times w^{o}_{k} \times h^{o}_{k}$. Thus, pruning filters not only reduces the number of computations in this layer but also reduces the size of the input to the next layer (which is the same as the output of this layer). The same process of pruning can then be repeated across all layers of the CNN. The main task here is to find the right $scoring\_function$ for ranking the filters.

\subsection{A Generic Algorithm for Pruning}
Algorithm \ref{algo} summarizes the generic recipe used by different approaches for pruning filters. As shown in the algo \ref{algo}, pruning typically starts from the outermost layer. Once the low scoring filters from this layer are pruned, the network is then fine-tuned and the same process is then repeated for the layers before it. Once all the layers are pruned and fine-tuned, the entire network is then tuned for a few epochs.

\begin{algorithm}
\caption{\textit{Prune(CNN)}}\label{algo}
\begin{algorithmic}[1]
\STATE $K \leftarrow$ number of layers in the network\\
%\STATE $Pruned\_CNN \leftarrow CNN$ (initialized to original trained network)\\
\STATE $\boldsymbol{F_k} = \{ F_{k1}, F_{k2}, ..., F_{kn} \}$ (filters in layer $k$)\\
\FOR{each layer $k\in K \dots 1$}

	\FOR{each filter $F_{ki} \in F_{k1}, F_{k2}, ..., F_{kn}$}
    
		\STATE $score_{ki} = scoring\_function(F_{ki})$
    \ENDFOR
    \STATE $\boldsymbol{F^{'}_{k}} = top\_m\_filters(F_k, score_{k1}, ..., score_{kn})$
    \STATE $CNN = retain\_filters(CNN, \boldsymbol{F^{'}_{k}}) $
    \STATE Finetune $CNN$ for $p$ epochs
\ENDFOR
\STATE Finetune the final pruned $CNN$ for $q$ epochs
\end{algorithmic}
\end{algorithm}

Existing methods for pruning filters differ in the $scoring\_function$ that they use for ranking the filters. We alternately refer to this $scoring\_function$ as pruning criteria as discussed in the next subsection.

\subsection{Pruning Criteria}\label{criteria}
We now describe various pruning criteria which are used by existing approaches and also introduce some new variants of existing pruning criteria. These criteria are essentially used as $scoring\_function()$ in Algorithm \ref{algo}.

\begin{enumerate}

\item Mean Activation \cite{efficientInference} : Most deep CNNs for image classification use ReLU as the activation function which results in very sparse activations (as all negative outputs are set to 0). We could compute the mean activation of the feature map corresponding to a filter across all images in the training data. If this mean activation is very low (because most of the activations are 0) then this feature map and hence the corresponding filter is not going to contribute much to the discriminatory power of the network (since the filter rarely fires for any input). Hence, \cite{efficientInference} uses the mean activation as a $scoring\_function$ for ranking filters. 

\item $l_{1}$-Norm \cite{l1norm} : The authors of \cite{l1norm} suggest that the $l_{1}$-norm  ($\parallel$F$\parallel_{1}$) of a filter can also be used as an indicator of the importance of the filter. The argument is that if the $l_{1}$-norm of a filter is small then on average the weights in the filter will be small and hence produce very small activations. These small activations will not influence the output of the network and hence the corresponding filters can be pruned away. One important benefit of this method is that apart from computing the $l_{1}$-norm, it does not need any extra computation during pruning and fine-tuning. 

\item Entropy \cite{entropy} : If the feature map corresponding to a filter produces the same output for every input (image) then this feature map and hence the corresponding filters may not be very important (because it does not play any discriminatory role). In other words, we are interested in feature maps (and hence filters) which are more informative or have a high entropy. If we divide the possible range of the average output of a feature map into $b$ bins then we could compute the entropy of the $i^{th}$ feature map (or filter) \cite{entropy} as :
\begin{align*}
E_{i} = -\sum_{j=1}^{b}p_{ij}\log p_{ij}     
\end{align*}

where $p_{ij}$ is the probability  that the output of the $i^{th}$ feature map lies in the $j^{th}$ bin. This probability can be computed as the fraction of  input images for which the average output of the feature map lies in this bin.

\item Average Percentage of Zeros (APoZ) \cite{apoz} : As mentioned earlier, when ReLU is used as the activation function, the output activations are very sparse. If most of the neurons in a feature map are zero then this feature map is not likely to contribute much to the output of the network. The Average Percentage of Zeros in the output of each filter can thus be used to compute the importance of the filter (the lesser the better). 

\item Sensitivity : We could compute the gradient of a filter w.r.t. the loss function (i.e, cross entropy). If a filter has a high influence on the loss function then the value of this gradient would be high. The $l_{1}$-norm of this gradient averaged over all images can thus be used to compute the importance of a filter. 

\item Scaled Entropy : We propose a new variant of the entropy based criteria. We observe that a filter may have a high entropy but if all its activations are very low (belonging to lower bins) then this filter is not likely to contribute much to the output. We thus propose to use a combination of entropy and mean activation by scaling the entropy by the mean activation of the filter. This scaled-entropy of $i^{th}$ filter can be computed as:\\
\begin{equation*}
SE_{i} = -\sum_{j=1}^{b}p_{ij}log p_{ij} * Mean_{i}
\end{equation*}
where $Mean_{i}$ is the average activation of the $i^{th}$ filter over all the input images. 

\item Class Specific Importance : In this work, we are also interested in a more practical scenario, where a network trained for detecting all the 1000 classes from ImageNet is required to detect only ($l < 1000$) of these classes at test time (say, only animals). Intuitively, we should then devise a $scoring\_function$ which retains only those filters which are important for these $l$ classes. To do so we compute the gradient of loss function w.r.t. the filter after the training is done. However, now instead of averaging the $l_{1}$-norm of this gradient over all images in the training data, we compute the average over only those images in the training data which correspond to the $l$ classes of interest. This class-specific average is then used to rank the filters.

\item Random Pruning : One of the main contributions of this work is to show that even if we randomly prune the filters from a CNN, its performance after fine-tuning is not much worse than any of the above approaches. 
\end{enumerate}

\section{Experiments: Image Classification}
In this section, we focus on the task of image classification using the ImageNet \cite{ILSVRC15} dataset. The dataset is split into three sets : training (1.3M images), validation (50K images), and testing (100K images with held-out class labels). We experiment with two popular networks, \textit{viz.}, VGG-16 and ResNet-50. We first train these networks using the full ImageNet training data and then prune them using Algorithm \ref{algo}. We compare the performance of different scoring functions as listed in the section \ref{criteria}. 

\subsection{Comparison of different pruning methods on VGG-16}
\begin{figure*}[ht]

    \begin{subfigure}[b]{0.5\textwidth}
        \includegraphics[width=\textwidth,height=0.20\textheight]{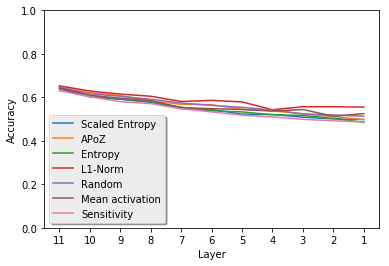}
        \caption{Performance after fine-tuning at each layer in VGG-16}
        \label{fig-recovery}
    \end{subfigure}
        \begin{subfigure}[b]{0.5\textwidth}
      \includegraphics[width=\textwidth,height=0.20\textheight]{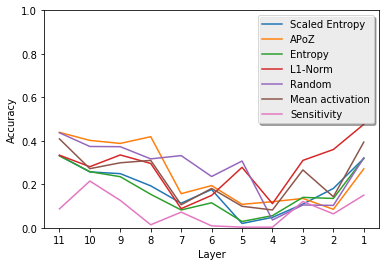}
      \caption{Performance drop after pruning but before fine-tuning}
      \label{fig-damage}
    \end{subfigure}
	\begin{subfigure}[b]{0.5\textwidth}
    \includegraphics[width=\textwidth,height=0.20\textheight]{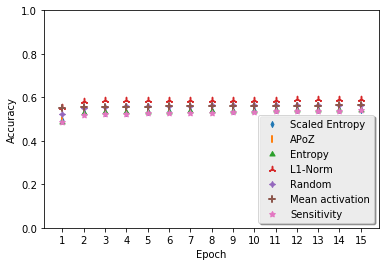}
    \caption{Finetuning final pruned network with 1/$10^{th}$ data}
    \label{fig-speed}
    \end{subfigure}
    \begin{subfigure}[b]{0.5\textwidth}
    \includegraphics[width=\textwidth,height=0.20\textheight]{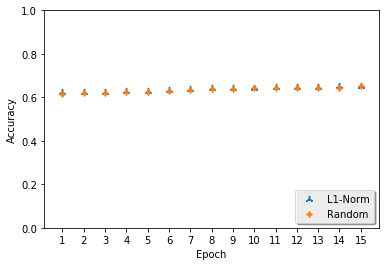}
    \caption{Finetuning final pruned network with full ImageNet data}
    \label{fig-full-data}
    \end{subfigure}

    \caption{Pruning and Fine-tuning VGG-16}\label{Images}
\end{figure*}

VGG-16 \cite{vgg16} has 13 convolutional (CONV) and two fully connected (FC) layers. The number of filters in each CONV layer in the the standard VGG-16 network \cite{vgg16} is \{64, 64, 128, 128, 256, 256, 256, 512, 512, 512, 512, 512, 512\}. We first train this network as it is (\textit{i.e.}, with the standard number of filters in each layer) using the ImageNet training data. When evaluated on the standard ImageNet test set, this trained model gives us a top-$1$ accuracy of 69.91$\%$ which is comparable to the accuracy reported elsewhere in the literature. We now prune this network, one layer at a time starting from the last convolution layer. We prune away $m$\% of filters from each layer where we chose the value of $m$ to be \{25, 50, 75\}. We use one of the scoring functions described in Section \ref{criteria} to select the top $m$\% filters. We drop the remaining (100 - m)\% filters from this layer and then fine-tune the pruned network for 1 epoch. We then repeat the same process for the lower layers and use the same value of $m$ across all layers. Once the network is pruned till layer 1, we then fine tune the entire pruned network for 12 epochs using 1/10-th of the training data picked randomly. The only reason for not using the entire training data is that it is quite computationally expensive. We did not see any improvement in the performance on the validation set by fine-tuning beyond 12 epochs. We then evaluate this pruned and fine-tuned network on the test set. Below, we discuss the performance of the final pruned and fine-tuned network obtained using different pruning strategies.\\
\noindent \textbf{Performance of pruned network after fine-tuning:} In Table \ref{tab-2}, we report the performance of the final pruned network after fine tuning. We observe that random pruning works better than most of the other pruning methods described earlier. $l_1$-norm is the only $scoring\_funct$-$ion$ which does better than random and that too by a small margin. In fact, if we fine-tune the final trained network using the entire training data then we observe that there is hardly any difference between random and $l_1$-norm (see Table \ref{tab-1}). This provides empirical evidence for our claim that the amount of recovery ($i.e$,  final performance after fine-tuning) is not due to the soundness of the pruning criteria. Even with random pruning, the performance of the pruned network is comparable. Of course, as the percentage of pruning increases ($i.e,$ as $m$ increases) it becomes harder for the pruned network to recover the full performance of the original network (but the point is that it is equally hard irrespective of the pruning method used). Thus, w.r.t. the amount of recovery after damage (pruning), a random pruning strategy is as good as any other pruning strategy. We further drive this point in Figure \ref{fig-recovery} where we show that after pruning and fine tuning for every layer, the amount of recovery after fine tuning is comparable across different pruning strategies.
\begin{table}
\centering
\begin{tabular}{llll}
\hline
Heuristic        & 25 \%    & 50\%     & 75\%    \\ \hline
Random          & 0.650  & 0.569  & 0.415 \\
Mean Activation & 0.652   & 0.570  & 0.409   \\
Entropy         & 0.641 & 0.549   & 0.405  \\
Scaled Entropy  & 0.637   & 0.550  & 0.401  \\
$l_1$-norm         & \textbf{0.667}   & \textbf{0.593}  & \textbf{0.436}  \\
APoZ            & 0.647 & 0.564 & 0.422 \\
Sensitivity     & 0.636  & 0.543  & 0.379 \\ \hline
\end{tabular}
\caption{Comparison of different filter pruning strategies on VGG-16.}
\label{tab-2}
\end{table}

As a side note we would like to mention that we do not include the performance of ThiNets \cite{thinet} in Table \ref{tab-2}. This is because it uses a slightly different methodology. In particular there are two major differences. First, in ThiNets pruning is done only till layer 10 and not upto layer 11 as is the case for all numbers reported in Table \ref{tab-2}. Secondly, in ThiNets, if a CONV layer appears before a max-pooling layer then it is fine-tuned for an extra epoch to compensate more for the downsampling in the max pooling layer. For a fair comparison, we followed this exact same strategy as ThiNet but using a random pruning criteria. In this setup, a randomly pruned network was able to achieve 68\% top-1 accuracy after 50\% pruning which is comparable to the performance of the corresponding ThiNet (69\%).

\begin{table}
\centering
\begin{tabular}{llll}
\hline
Heuristic         & 50\%    \\ \hline
Random          & 0.6701 \\
Mean Activation & 0.6662 \\
Entropy         & 0.6635 \\
Scaled Entropy  & 0.6625 \\
\textbf{$l_1$-norm}      & \textbf{0.6759} \\
APoZ            & 0.6706 \\
Sensitivity     & 0.6659 \\ \hline
\end{tabular}
\caption{Performance after fine-tuning with full data}
\label{tab-1}
\end{table}

\noindent \textbf{Amount of initial damage caused by different pruning strategies:} One might argue that while random pruning strategy is equivalent to other pruning strategies w.r.t. final performance after fine tuning, it is possible that the amount of initial damage caused by a careful pruning strategy maybe less than than caused by random pruning. This could be important in cases where enough time or resources are not available for fine-tuning after pruning. To evaluate this, we compute the accuracy of the network just after pruning (and before fine-tuning) at each layer. Figure \ref{fig-damage} compares this performance for different pruning strategies. Here again we observe that the damage caused by a random pruning strategy is not worse than other pruning strategies. The only exception is when we prune the first 4 layers in which case the damage caused by $l_1$-norm based pruning is less than random pruning. We hypothesize that this is because the first 4 layers have very few filters and hence one needs to be careful while pruning for filters from these layers. In fact, in hindsight we would recommend not to prune any filters from these 4 layers because the computation savings are less as compared to drop in accuracy. 

\noindent\textbf{Speed of recovery and quantum of data for fine-tuning:}
Another important criteria is the speed of recovery, \textit{i.e.}, the number of iterations for which the network needs to be fine-tuned after pruning. It is conceivable that a carefully pruned network may be able to recover and reach its best performance faster than a randomly pruned network. However, as shown in Figure \ref{fig-speed} that almost all the pruning strategies (including random) reach their peak after 2 epochs when fine-tuned with one-tenth of the data. Even, if we increase the quantum of data, this behavior does not change as shown in Figure \ref{fig-full-data} (for $l_1$-norm based pruning and random pruning). Of course, as we increase the quantum of data the amount of recovery increases, \textit{i.e.}, the peak performance of the pruned network increases. However, the important point is that a random strategy is no worse than a careful pruning strategy w.r.t. speed of recovery and quantum of data required. 

\subsection{Which layers to pre-train?}
So far in all our experiments, we have re-trained all the layers after pruning a given layer. We wanted to check if this is indeed necessary or it possible to re-train only a few layers. To assess this we explore various choices as described below: 

\noindent \textbf{Retraining only fully connected layers:} The fully connected layers have many parameters and should perhaps be able to adjust to any perturbations such as pruning in the initial layers To verify this hypothesis, we ran experiments in which we only re-trained the fully connected layer after pruning filters from a given layer. Specifically, we do not retrain any of the convolution layers. In this case, the final top-1 accuracy we got is 8\% as compared to 67\% which we got when we retrained both fully connected layers and convolutional layers. An intuitive explanation for this is that the convolutional layers which act as feature extractors are as important as the set of fully connected layers. In particular, since in every iteration, we are dropping more filters and not retraining the convolutional layers the resulting feature vector of the image at the end of last convolutional layer is so bad that despite having a large number of parameters, the fully connected layers are not being able to deal with this bad representation.

\noindent \textbf{Retraining only convolutional layers:} Next, for the sake of completeness, we decided to see what happens if we retrain only the convolution layers and not the fully connected layers. Surprisingly, we observed that in this case we get a top-1 accuracy of 66.23\% as compared to 67\% that we get after training both convolutional and fully connected layers.  This suggests that the convolutional layers are able to adapt and produce a good representation which is compatible with a fixed fully connected layer.

\noindent \textbf{Retraining only neighboring layers:} Further, we wanted to check what happens if we retrain only the neighboring layers of the pruned connected layers. For example, if we prune layer $i$ then we retrain only layers $i-1$, $i$ and $i+1$. The intuition here was that after disrupting one layer, it is enough to simply ensure that its neighboring layers adjust to this disruption. In this case, we get top-1 accuracy of 57.9\% which is very low as compared to the top-1 accuracy of 67\% which we get after retraining all the layers in the network. This is due to a cascade effect wherein once layer $i+1$ adjusts to the pruning in layer $i$, layer $i+2$ also needs to adjust to the changes in the layer $i+1$ and so on. However since we are not retraining subsequent layers, so the overall performance drops.

\subsection{Training Smaller Network from Scratch}
Table \ref{tab-1} suggests that we can prune a pre-trained network to half of its original size and still get the same performance as that obtained by the original network. An obvious question here is that instead of training a bigger network and then pruning it to half of its original size is it possible to train a smaller network from scratch whose size/architecture is the same as the final pruned network. To verify this, we trained a smaller VGG-16 network (VGG-16/2) in which we reduced the number of filters in all the convolutional layers by 50\%. For example the number of filters in each CONV layer in the the standard VGG-16 network \cite{vgg16} is \{64, 64, 128, 128, 256, 256, 256, 512, 512, 512, 512, 512, 512\}. Now in VGG-16/2 the number of filters in each CONV layer is \{32, 32, 64, 64, 128, 128, 128, 256, 256, 256, 256, 256, 256\}. In other words, the architecture of VGG-16/2 is the same as that that obtained after pruning 50\% filters from each convolution layer of VGG-16 using Algorithm \ref{algo}.

\begin{figure}[h]
    \centering
    \includegraphics[scale=0.55]{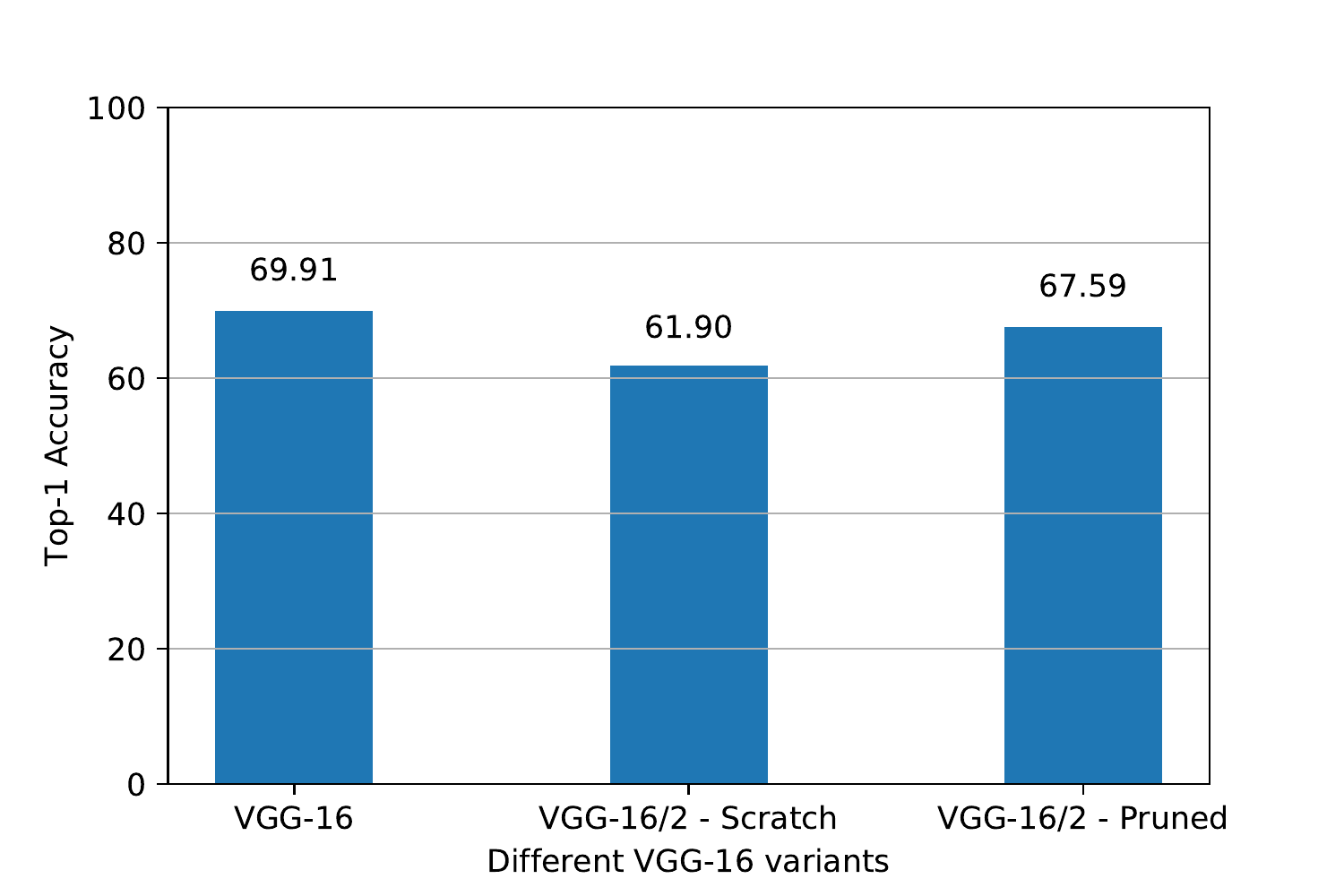}
    \caption{Performance drop after pruning but before fine-tuning}
    \label{fig-scratch_vs_prune}
\end{figure}

Figure \ref{fig-scratch_vs_prune} compares the top-1 accuracy of VGG-16 (first bar) with that of VGG-16/2 (second bar) and that obtained by pruning 50\% of the filters from a trained VGG-16 model using $l_1$-norm criterion and then fine-tuning the pruned network (third bar). The main observation here is that VGG-16/2, when trained from scratch, achieves 61.90\% top-1 accuracy which is very less as compared to the accuracy achieved by the pruned network. This is an interesting result which suggests that it is beneficial to first train a bigger network and then prune out filters from it instead of directly training a smaller network from scratch.

\subsection{Random Pruning v/s $l_1$ norm: Some more insights}
\begin{figure*}[t]

    \begin{subfigure}[]{0.5\textwidth}
        \includegraphics[width=\textwidth,height=0.20\textheight]{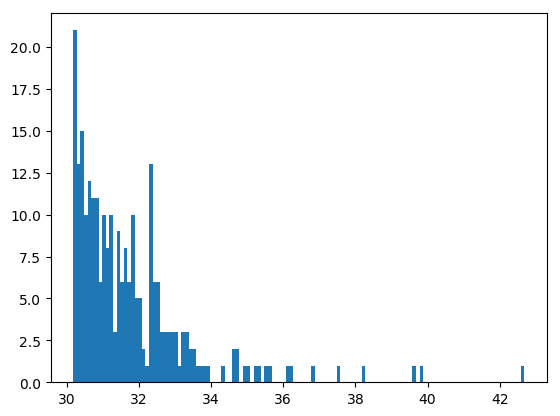}
        \caption{Distribution of the top 50\% filters selected using the $l_1$-norm}
        \label{fig-recovery-1}
    \end{subfigure}\begin{subfigure}[]{0.5\textwidth}
      \includegraphics[width=\textwidth,height=0.20\textheight]{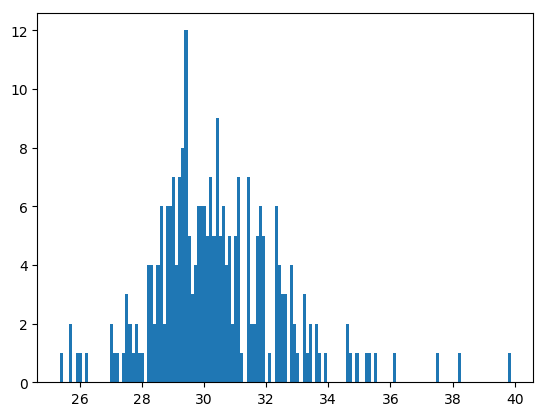}
      \caption{Distribution of the top 50\% filters selected randomly.}
      \label{fig-damage-1}
    \end{subfigure}
	\begin{subfigure}[]{0.5\textwidth}
    \includegraphics[width=\textwidth,height=0.20\textheight]{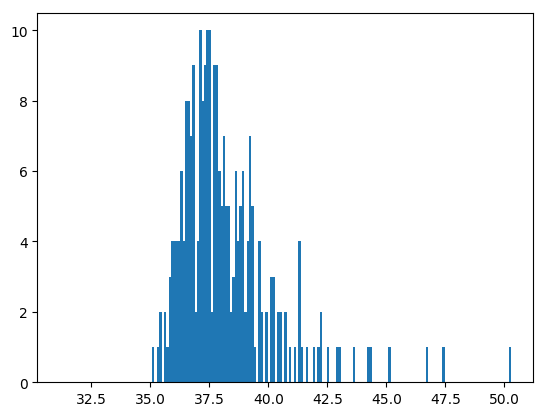}
    \caption{Distribution of the corresponding finetuned model in (a)}
    \label{fig-speed-1}
    \end{subfigure}\begin{subfigure}[]{0.5\textwidth}
    \includegraphics[width=\textwidth,height=0.20\textheight]{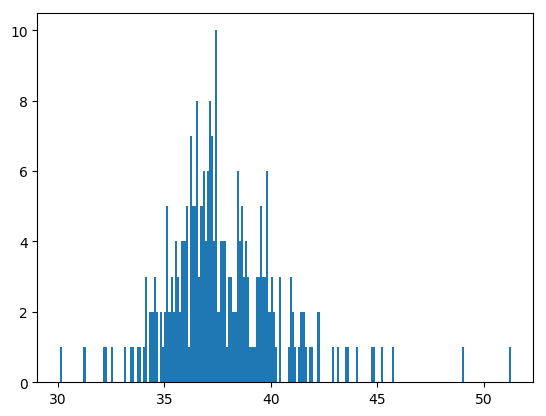}
    \caption{Distribution of the corresponding finetuned model in (b)}
    \label{fig-full-data-1}
    \end{subfigure}
\caption{Distribution of $l_1$-norm values of filters in $11^{th}$ layer of VGG-16.  \textit{x-axis} is the range of $l_1$-norm values and \textit{y-axis} denotes the number of filters.}
\label{fig:rand-explain-11}
\end{figure*}

In Figure \ref{fig:rand-explain-11}, we have shown the distribution of $l_1$-norms of filters in the $11^{th}$ layer of VGG-16. For all the histograms in Figure \ref{fig:rand-explain-11}, \textit{x-axis} is the range of $l_1$-norm values and \textit{y-axis} denotes the number of filters. Figure \ref{fig:rand-explain-11}(a) shows the distribution of the top 50\% filters selected using the $l_1$-norm criteria. As expected this distribution is highly skewed. However, on retraining this pruned network, this skewed distribution again moves towards a normal distribution as shown in Figure \ref{fig:rand-explain-11}(c). On the other hand, when we randomly select 50\% filters, the $l_1$-norms of these filters follow a normal distribution as shown in Figure \ref{fig:rand-explain-11}(b). Even after retraining, the $l_1$-norms of these filters are normally distributed as shown in Figure \ref{fig:rand-explain-11}(d). This behavior indicates that $l_1$-norm is perhaps not the best criteria as the network eventually prefers filters such that their $l_1$-norms are normally distributed.

\begin{table}[]
\begin{tabular}{cccc}
\hline
Layer   & Original & 50\% Pruning & Differential\\
Number  &          &              & Pruning \\
\hline
1             & 64       & 32           & 59          \\
2             & 64       & 32           & 59          \\
3             & 128      & 64           & 108         \\
4             & 128      & 64           & 108         \\
5             & 256      & 128          & 175         \\
6             & 256      & 128          & 175         \\
7             & 256      & 128          & 175         \\
8             & 512      & 256          & 185         \\
9             & 512      & 256          & 185         \\
10            & 512      & 256          & 185         \\
11            & 512      & 256          & 185         \\
\hline
\textbf{Total Filters} & 3200     & 1600         & 1599 \\       
\hline
\end{tabular}
\caption{Pruning filters in ratio of the number of filters present in the original model.}
\label{tab:ratio-random}
\end{table}

\subsection{Differential Pruning based on number of filters in a layer}
So far in all our experiments, we have pruned the same percentage of filters from all the layers. However, this is a bit unfair as the lower layers have very few filters as compared to the upper layers. For example, $layer 1$ has only 64 filters whereas $layer$10 has 512 filters. Intuitively, it seems that pruning out 32/64 filters is more brutal then pruning 256/512 filters. More so because disruptions at the lower layers are likely to cascade through the entire network. To prove/disprove this hypothesis, we ran an experiment where we prune filters in proportion to the total number of filters in a layer. More specifically, the first eleven layers of VGG-16 has a total of 3200 filters of which we want to prune out 50\% of the filters (i.e, prune away a total of 1600 filters). However, now instead of pruning 50\% of the filters from each layer we prune fewer filters from layers which have fewer filters while ensuring that the total number of filters pruned is still 1600. In other words, the filters are pruned in proportion to the number of filters in that layer. In Table \ref{tab:ratio-random}, Differential-Pruning column shows the number of filters retained in each layer when pruning according to the scheme described above. In particular, note that more number of filters are pruned from higher layers. However, the final results suggest that there is not much improvement by using this differential pruning strategy wherein the top-1 accuracy improves to $0.6758$ top-1 accuracy as compared to $0.6701$ earlier.

\subsection{Pruning by averaging}
One way of reducing the size of the network by 50\% is to take the average of consecutive filters and then dropping both the filters. For example, say in layer 10 we have 512 filters, then on taking the average of consecutive filters we will be left with 256 filters.  We wanted to check if this would be better than just dropping one of the filters, because now the average at least encodes the original information available in both the filters. When we pruned VGG-16 (by 50\%) using the strategy and then retrained it we were able to achieve a top-1 accuracy of 0.6576 which was 0.1250 less as compared to the case when we randomly drop 50\% of the filters instead of averaging. This suggests that averaging the filters does more harm as the original information from both the filters gets morphed when we take an average. 

\subsection{Pruning ResNet-50 using $l_1$-Norm and Random}
While the above set of experiments focused on VGG-16, we now turn our attention to ResNet-50 \cite{resnet} which gives state-of-the-art results on ImageNet. We took a trained ResNet-50 model which gave 74.5$\%$ top-1 accuracy on the ImageNet test set which is again comparable to the accuracy reported elsewhere in the literature. ResNet contains 16 residual blocks wherein each block contains 3 layers with a skip connection from the first layer to the third layer. The standard practice is to either prune the first layer of each block or the first two layers of each block. In the first case, out of the total 48 convolution layers (16 * 3) we will end up pruning 16 and in the second case we will end up pruning 32 layers. As before, for each pruned layer we vary the percentage of pruning from 25\%, 50\% to 75\%. Here, we only compare the performance of $l_1$-Norm with random pruning as these were the top performing strategies on VGG-16. This was just to save time and resources as given the deep structure of ResNet it would have been very expensive to run all pruning strategies. Once again from Table \ref{tab-3}, we observe that \textbf{random} pruning performs at par (in fact, slightly better) when compared to $l_{1}$-Norm based pruning. Note that, in this case the pruned models were trained with only one-tenth of the data. The performance of both the methods are likely to improve further if we were to fine-tune the pruned network on the entire training data. 
\begin{table}[h!]
\centering
\begin{tabular}{lclll}
\hline
Heuristics & \multicolumn{1}{l}{\#Layers Pruned} & 25 \%  & 50\%   & 75\%   \\ \hline
Random     & 16                                  & 0.722 & 0.683 & 0.617 \\
$l_1$-norm    & 16                                  & 0.714 & 0.677 & 0.610 \\
Random     & 32                                  & 0.696 & 0.637 & 0.518 \\
$l_1$-norm    & 32                                  & 0.691 & 0.633 & 0.514 \\ \hline
\end{tabular}
\caption{Comparison of different filter pruning strategies on ResNet (Top-1 accuracy of unpruned network is 0.745)}
\label{tab-3}
\end{table}
%These experiments make the notion of `Random pruning performing well' more stronger.

\section{Experiments: Class specific pruning}
Existing work on pruning filters (or model compression, in general) focuses on the scenario where we have a network trained for detecting all the 1000 classes in ImageNet and at test time it is again evaluated using data belonging to all of these 1000 classes. However, in many real world scenarios at test time we may be interested in fewer classes. A case in point, is the Pascal VOC dataset which contains only 20 classes. Intuitively, if we are interested in only fewer classes at test time then we should be able to prune the network to cater to only these classes. Alternatively, we could train the original network itself using data corresponding to these classes only. To enable these experiments, we first create a new benchmark from ImageNet which contains only those 52 classes which correspond to the 20 classes in Pascal VOC. Note that the mapping of 52-20 happens because ImageNet has more fine-grained classes. For example, there is only one class for `dog' in Pascal VOC but ImageNet contains many sub-classes of `dog' (different breeds of dogs). We manually went over all the classes in ImageNet and picked out the classes which correspond to the 20 classes in Pascal VOC. In some cases, we ignored ImageNet classes which were too fine-grained and only considered those classes which were immediate hyponyms of a class in Pascal VOC. We then extracted the train, test and valid images for these classes from the original ImageNet dataset. We refer to this subset of ImageNet as ImageNet-52P (where P stands for Pascal VOC). We refer to the original ImageNet dataset as ImageNet-1000. Note that the train, test and validation splits of ImageNet-52P are subsets of the corresponding splits of ImageNet-1000. In particular , the training split of ImageNet-1000 does not overlap with the test or validation splits of ImageNet-52P.

We first compare the performance in the following two setups: (i) model trained on ImageNet-1000 and evaluated on the test split of ImageNet-52P and (ii) model trained on ImageNet-52P and evaluated on the test split of ImageNet-52P. We observe that while in the first setup we get a top-1 accuracy of 74\%, in the second setup we get an accuracy of 87\%. This suggests that model trained on ImageNet-1000 is clearly overloaded with extra information about the remaining 948 classes and hence performs poorly on the 52 classes of interest. We should thus be able to prune the network effectively to cater to only the 52 classes of interest. Note that in practice it is desirable to have just one network trained on ImageNet-1000 and then prune it for different subsets of classes that we are interested in instead of training a separate network from scratch for each of these subsets. We again compare different pruning strategies as listed earlier except that now when fine-tuning (after each layer and at the end of all layers) we only use ImageNet-52P. In other words, we fine-tune using only data corresponding to the 52 classes. Once again, we observe that there is not much difference between random pruning and other pruning strategies. Also with 25\% pruning, we are almost able to match the performance of a network trained only on these 52 classes (\textit{i.e}, 87\%). 

\begin{table}
\centering
\begin{tabular}{llll}
\hline
Heuristics        & 25 \%    & 50\%     & 75\%    \\ \hline
Random            & 0.859  & 0.820   & 0.692  \\ 
Mean Activation   & 0.866  & 0.816   & 0.698   \\ 
Entropy           & 0.860  & 0.802   & 0.684  \\ 
Scaled Entropy    & 0.863  & 0.813   & 0.691  \\ 
$l_1$-norm           & \textbf{0.867}  & \textbf{0.823}   & \textbf{0.729}  \\ 
APoZ              & 0.858 & 0.811    & 0.700     \\ 
Important Classes & 0.857   & 0.795    & 0.655   \\ 
Sensitivity       & 0.849  & 0.793    & 0.634 \\ \hline
\end{tabular}
\caption{Comparison of different filter pruning strategies when fine-tuned and evaluated with ImageNet-52P.}
\label{tab-6}
\end{table}

\section{Experiments: Faster Object Detection}

The above experiments have shown that with reasonable levels of pruning (25-50\%) and enough fine-tuning (using entire data) the pruned network is able to recover and almost match the performance of the unpruned network on the original task (image classification) even with a random pruning strategy. However, it is possible that if such a pruned network is used for a new task, say object detection, then a randomly pruned network may not give the same performance as a carefully pruned network. To check this, we perform experiments using the Faster-RCNN model for object detection. Note that the Faster-RCNN model uses a VGG-16 model as a base component and then adds other components which are specific to object detection. We experiment with the PASCAL-VOC 2007 dataset \cite{pascal-voc-2007} which consists of 9,963 images, containing 24,640 annotated objects. We first plug-in a standard trained VGG-16 network into Faster-RCNN and then train Faster-RCNN for 70K iterations (as is the standard practice). This model gives a mean Average Precision (mAP) value of $0.66$. The idea is to now plug-in a pruned VGG-16 model into faster RCNN instead of the original unpruned model and check the performance. Table \ref{tab-4} again shows that the specific choice of pruning strategy does not have much impact on the final performance on object detection. Of course, as earlier, as the level of pruning increases the performance drops (but the drop is consistent across all pruning strategies). We now report some more interesting experiments on pruning Faster RCNN.

\begin{table}
\centering
\begin{tabular}{llll}
\hline
Heuristics        & 25 \%    & 50\%     & 75\%    \\ \hline
Random          & \textbf{0.647} & 0.600 & 0.505 \\
Mean Activation & \textbf{0.647} & 0.601    & 0.489 \\
Entropy         & 0.635 & 0.584  & 0.501 \\
Scaled Entropy  & 0.640  & 0.593  & 0.507 \\
$l_1$-norm         & 0.628 & \textbf{0.608}  & \textbf{0.520}  \\
APoZ            & 0.646 & 0.598  & 0.514 \\
Sensitivity     & 0.636 & 0.592  & 0.485\\ \hline
\end{tabular}
\caption{Object detection results obtained by plugging-in different pruned VGG-16 models into Faster-RCNN.}
\label{tab-4}
\end{table}

\begin{table}[H]
\centering
\begin{tabular}{l|llll}
\hline
Faster-RCNN        & Baseline & 25 \%    & 50\%    & 75\%    \\ \hline
mAP  & 0.66 & 0.655 & 0.648 & 0.530 \\ \hline
fps  & 7.5	& 10  & 13	& 16 \\ \hline
\end{tabular}
\caption{Object detection results when directly pruning (random) a fully trained Faster-RCNN model.}
\label{tab-5}
\end{table}

\subsection{Directly pruning Faster RCNN}
Instead of plugging in a pruned VGG-16 model into Faster-RCNN, we could alternatively take a trained Fas-ter RCNN model and then prune it directly. Here again, we use a simple random pruning strategy and observe that the performance of the pruned model comes very close to that of the unpruned model. In particular, with 50\% pruning we are able to achieve a mAP of $\textbf{0.648}$ with a $74\%$ speedup in terms of frames per second (Table \ref{tab-5}).

\begin{table}
\centering
\begin{tabular}{llll}
\hline
Heuristics        & 25 \%    & 50\%     & 75\%    \\ \hline
Random            & 0.647 & 0.580 & 0.469  \\
Mean Activation   & 0.644 & 0.583 & 0.454  \\
Entropy           & 0.642 & 0.578 & 0.470   \\
Scaled Entropy    & 0.645 & 0.580 & 0.443  \\
$l_1$-norm           & \textbf{0.648} & \textbf{0.601} & \textbf{0.487}  \\
APoZ              & 0.641 & 0.585 & 0.466  \\
Important Classes & 0.631 & 0.568 & 0.432  \\
Sensitivity       & 0.637 & 0.576 & 0.435 \\ \hline
\end{tabular}
\caption{Object detection results obtained by plugging-in different pruned VGG-16 models fine-tuned with ImageNet-52P as opposed to ImageNet-1000.}
\label{tab-7}
\end{table}

\subsection{Plugging in a VGG-16 model trained using Image-Net-52P}
\vspace{-2mm}
Since we are only interested in the 52 classes corresponding to Pascal-VOC, we wanted to check what happens if we plug-in a VGG-16 model trained, pruned and fine-tuned only on ImageNet-52P. As shown in Table \ref{tab-7} we do not get much benefit of plugging in this specialized model into Faster-RCNN. In fact, in a separate experiment we observed that even if we train a VGG-16 model on a completely random set of 52 classes (different from the 52 classes corresponding to Pascal VOC) and then plug in this model into Faster RCNN, even then the final performance of the Faster RCNN model remains the same. This is indeed surprising and further demonstrates the ability of these networks to recover from unfavorable situations.  

\begin{table}[H]
\centering
\begin{tabular}{lllll}
\hline
Heuristics                   &     Dataset          & 25 \%    & 50\%     & 75\%    \\\hline
\multirow{3}{*}{Random} & ImageNet-52R  & 0.650 & 0.590 & 0.463 \\
                        & ImageNet-52P  & 0.647 & 0.580 & 0.469 \\
                        & ImageNet-1000 & 0.647 & 0.602 & 0.505 \\
                        \hline
\multirow{3}{*}{$l_1$-norm}   & ImageNet-52R  & 0.650 & 0.603 & 0.485 \\
                        & ImageNet-52P  & 0.648 & 0.600 & 0.487 \\
                        & ImageNet-1000 & 0.628 & 0.608 & 0.520 \\\hline
\end{tabular}
\caption{Faster-RCNN using VGG pruned by selecting classes other than corresponding to Pascal VOC at random.}
\label{tab:random-classes}
\end{table}

\vspace{-10mm}
\subsection{Plugging in a VGG-16 model trained using random set of 52 classes}
\vspace{-2mm}
In a separate experiment we observed that even if we train a VGG-16 model on a completely random set of 52 classes referred as ImageNet-52R (different from the 52 classes corresponding to Pascal VOC) and then plug in this model into Faster RCNN, even then the final performance of the Faster RCNN model remains the same. In Table \ref{tab:random-classes} we have shown results of this experiment. This is indeed surprising and further demonstrates the ability of these networks to recover from unfavorable situations.

\if false
\begin{table}
\centering
\begin{tabular}{lllll}
\hline
Dataset & Heuristics        & 25 \%    & 50\%     & 75\%    \\ \hline
Dataset-1 & Random   & 0.51 & 0.50 & 0.38 \\
Dataset-1 & $l_1$-norm  & 0.52 & 0.49 & 0.40 \\
Dataset-2 & Random   & 0.52 & 0.48 & 0.37 \\
Dataset-2 & $l_1$-norm  & 0.54 & 0.49 & 0.39 \\ \hline
\end{tabular}
\caption{Faster RCNN results using pruned VGG-16 using Dataset-1 and Dataset-2, and not training VGG layers in Faster RCNN.}
\label{tab-8}
\end{table}
\fi 

\section{Experiments: Image Segmentation}
Lastly, we performed experiments with image segmentation where the goal is to assign each pixel to one of the given classes. We chose SegNet \cite{segnet} as our base model. SegNet has a convolutional encoder-decoder architecture. The encoder consists of a set of convolutional layers which compute a representation for the input. The decoder again contains a set of corresponding convolutional layers which upsample the low dimensional representation computed by the encoder to produce an output which is of the same size as the original image. The output from the final layer of the decoder is then passed to a pixel-wise classification layer. Out of the different variants of SegNet architecture, we used SegNet-Basic (as described in \cite{segnet}). which contains 8 convolutional (CONV) layers of which the first 4 act as encoder and the last four act as decoder. The number of filters in each CONV layer is 64. When trained this network using the the CamVid \cite{camvid} dataset and achieved a baseline accuracy of $0.78$.

We tried different experiments wherein in some experiments we only pruned filters from the decoder and in some experiments we pruned filters from both the encoder and the decoder. The rational here is that the effects of pruning the encoder could cascade to the decoder also and hence it makes sense to keep the encoder as it is and prune only the decoder. The results of our experiments are summarized in Table \ref{tab:segnet} where we compare random pruning with $l_1$-norm based pruning. We observe that random pruning clearly outperforms $l_1$-norm based pruning when we pruning both the encoder and the decoder. However, if we only prune the decoder, then both random pruning and $l_1$-norm strategy outperform the baseline results (full network) and random pruning giving the best result in case of 50\% pruning. We hypothesize that this better performance is due to the regularization effect of pruning.

\begin{table}[]
\begin{tabular}{cccc}
\hline
                      & Pruning(\%) & Decoder & Decoder+Encoder \\
\hline
\multirow{3}{*}{Random} & 25                    & 0.7894  & 0.7701  \\
                      & 50                    & 0.7948  & 0.7678  \\
                      & 75                    & 0.7899  & 0.5944  \\
\hline
\multirow{3}{*}{$l_1$-norm} & 25                    & 0.7825  & 0.6538  \\
                      & 50                    & 0.7813  & 0.3992  \\
                      & 75                    & 0.7891  & 0.3415 \\
\hline
\end{tabular}
\caption{Pruning results on SegNet-Basic architecture.}
\label{tab:segnet}
\end{table}

\section{Conclusion and Future Work}
We evaluated the performance of various pruning strategies based on the (i) drop in performance after pruning (ii) amount of recovery after pruning (iii) speed of recovery and (iv) amount of data required. We did extensive evaluations with two networks (VGG-16 and ResNet50) and presented counter-intuitive results which show that w.r.t. all these factors a random pruning strategy performs at par with principled pruning strategies. We also showed that even when such a randomly pruned network is used for a completely new task it performs well. Next, we experimented with the task of object detection and showed that by randomly pruning filters from Faster RCNN we can get a 74\% speed-up w.r.t frames per second with only a 1\% drop in the performance. Lastly, we experimented with the task of image segmentation and showed that when we prune filters from only the decoder layer of SegNet we get some sort of a regularization network as a result of which the performance of the pruned network is better than that of the original unpruned network.

There are various possible future directions to this work. Given random pruning of filters has worked so well in convolutional neural networks, it is worth exploring this idea further in the case of recurrent neural networks. Another possible line of work is to establish theoretical grounds which explains the behavior of random pruning better. Designing more efficient training schedules by incorporating random pruning is another interesting area to explore.

\begin{acknowledgements}
We thank the Robert Bosch Centre for Data Science and AI (RBC-DSAI) and Intel India for supporting this research.
\end{acknowledgements}

\color{black}

\bibliographystyle{spmpsci}      

\bibliography{egbib}   % name your BibTeX data base

\end{document}